# A Workflow to Efficiently Generate Dense Tissue "Ground Truth" Masks for Digital Breast Tomosynthesis

Tamerlan Mustafaev[1], Oleg Kruglov[1], Margarita Zuley[1], Luana de Méro Omena[2], Guilherme Muniz de Oliveira[2], Vitor de Sousa França[2], Bruno Barufaldi[3], Robert Nishikawa[1] and Juhun Lee[1,4]

[1]Department of Radiology, University of Pittsburgh, Pittsburgh, Pennsylvania, USA
[2]Center of Informatics, Federal University of Paraíba, Paraiba, BRA
[3]Department of Radiology, University of Pennsylvania, Pennsylvania, USA
[4]Department of Industrial Engineering, University of Pittsburgh, Pittsburgh, Pennsylvania, USA

## Abstract:

**Background**

Digital breast tomosynthesis (DBT) is now the standard of care for breast cancer screening in the USA. Accurate segmentation of fibroglandular tissue in DBT images is essential for personalized risk estimation, but algorithm development is limited by scarce human-delineated training data.

In this study we introduce a time- and labor-saving framework to generate a human-annotated binary segmentation mask for dense tissue in DBT.

Our framework enables a user to outline a rough region of interest (ROI) enclosing dense tissue on the central reconstructed slice of a DBT volume and select a segmentation threshold to generate the dense tissue mask. The algorithm then projects the ROI to the remaining slices and iteratively adjusts slice-specific thresholds to maintain consistent dense tissue delineation across the DBT volume. By requiring annotation only on the central slice, the framework substantially reduces annotation time and labor. We used 44 DBT volumes from the DBTex dataset for evaluation. Inter-reader agreement was assessed by computing patient-wise Dice similarity coefficients between segmentation masks produced by two radiologists, yielding a median of 0.84. Accuracy of the proposed method was evaluated by having a radiologist manually segment the 20th and 80th percentile slices from each volume (CC and MLO views; 176 slices total) and calculate Dice scores between the manual and proposed segmentations, yielding a median of 0.83.

**Keywords**: *Image Processing, DBT, Process Automation, Healthcare Artificial Intelligence, Breast Cancer*

# 1. Background

Digital breast tomosynthesis (DBT) is increasingly used for breast cancer screening in the US. It provides quasi three-dimensional (3D) breast tissue characterization and can improve lesion detection compared with conventional 2D mammography [1], [2]. High breast density is associated with increased breast cancer risk [3], [4].

Lack of publicly available 3D dense tissue annotation datasets limits algorithm development. This is, in part, due to the time-consuming effort needed to delineate the dense tissue in every slice of the DBT volume, which can exceed 100 slices per single volume.

In this technical note, we present a time- and labor-saving framework that streamlines the process of dense tissue mask generation across DBT slices.

Our framework let a user to outline the rough region of interest (ROI) enclosing dense tissue in the central reconstructed slices of a DBT volume. Then the user selects a segmentation threshold for the dense tissue mask in that slice. The algorithm in our framework then projects the central slice ROI to remaining slices. The threshold values for those slices are iteratively adjusted to ensure the same amount of dense tissue captured across DBT slices. This procedure allows a user to annotate the central slice only such that we can reduce the time and labor of annotation significantly. To increase the usability of our framework, we developed a graphical user interface (GUI) and made it publicly available.

Two radiologists used our proposed framework to generate dense tissue masks for selected DBT samples (176 volumes from 44 patients) from the publicly available DBTex dataset [5]. To show the feasibility of using our method for generating ground truth masks for breast dense tissue in DBT, we evaluated inter-reader agreement between two radiologists by computing the Dice similarity scores between their dense tissue masks generated by our framework. In addition, segmentation accuracy was assessed by comparison with time-consuming conventional manual slice-by-slice segmentations, which served as the reference standard.

# 2. Methods

## 2.1 Dataset

This study utilized the DBTex dataset [5] which includes DBT volumes of 5,060 patients acquired by Hologic systems, to develop and evaluate the proposed time and labor saving framework. We randomly selected 176 "for-presentation" views (88 R/L CC and 88 R/L MLO) of 44 patients with BI-RADS 1 (normal) assessment for this study.

## 2.2 Proposed time- and labor-saving framework for generating breast dense tissue masks

The core idea of our framework is to let a user annotate a dense tissue mask on the central slice only, then our internal algorithm propagates the user input from the central slice to remaining slices in a DBT volume. A user, e.g., radiologist, first places a rough ROI enclosing dense tissue on the central slice and segment dense tissue assigning a threshold on the same slice. Our internal algorithm then projects the ROI on the remaining slices and updates the corresponding thresholds iteratively to ensure the same amount of dense tissue captured across DBT slices. This ROI-based and iterative slice-wise thresholding algorithm can account for intensity variability introduced by out-of-plane artifacts and edge blurring near the top and bottom slices. As the user only performs segmentation on the central slice only, our framework can save time and labor of segmentation significantly compared to full slice-by-slice manual segmentation. The detailed process involves the following steps (see Figure 1 for visual assistance):

1) <u>Load the DBT volume</u>. The user imports the complete DBT series through the GUI, which automatically displays the central slice for initial interaction (Fig. 1a).
2) <u>Define the dense-tissue region</u>. Using a polygon-drawing tool, the user delineates the fibroglandular area on the central slice (Fig. 1b). The GUI enables fine-tuning of the contour and excludes irrelevant regions such as the skin fold, nipple, pectoral muscle, or lymph nodes.
3) <u>Normalize image intensities</u>. The pixel values in each slice are normalized to a [0-1] range to ensure consistent thresholding across the stack in the subsequent steps.
4) <u>Interactive threshold selection</u>. Within the GUI (Fig. 1c), the user adjusts a slider to select the threshold that best separates dense from fatty tissue inside the polygon mask. The system provides real-time visual feedback and quantitative readouts of the segmented dense area on the central slice (i.e., the measurements). Once confirmed, this initial threshold serves as the reference for subsequent slices.
5) <u>Iterative propagation and optimization</u>. The algorithm propagates the polygon mask and thresholds the data iteratively to all slices in the DBT volume (Fig. 1d). For each slice, an automated search determines the threshold that yields a segmented dense-tissue area most consistent with the reference area from the central slice. The resulting 3D mask represents the volumetric distribution of dense tissue (Fig. 1e).

To facilitate this process, all the steps are done in an interactive graphical user interface (GUI) for creating breast density "ground truth" masks. The GUI streamlines the density mask generation process and provides a user-friendly interface. [1]

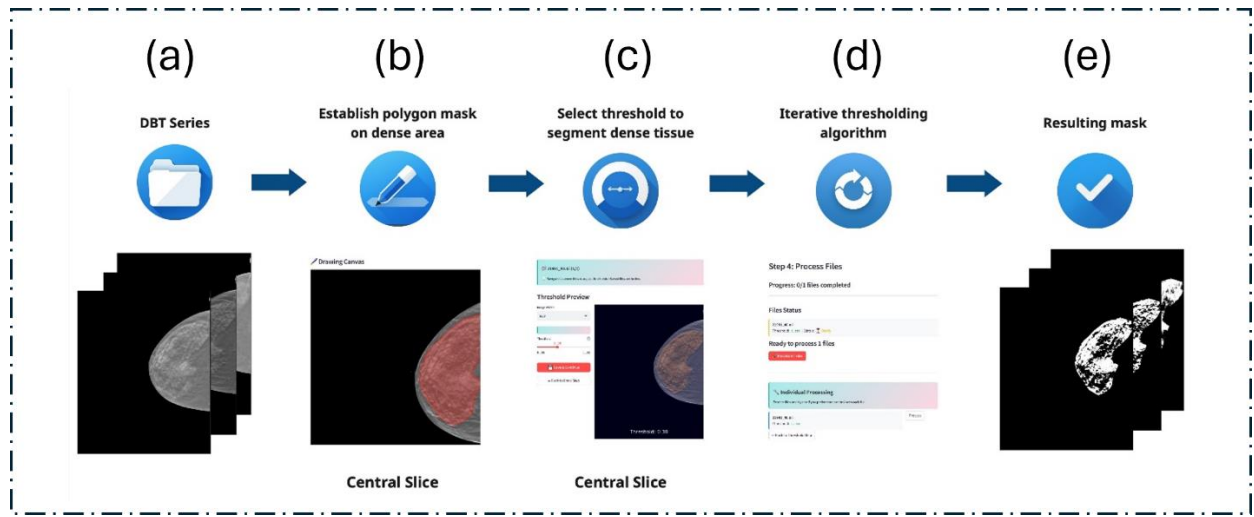

**Figure 1.** Proposed workflow for dense tissue mask generation. A GUI was developed to implement this workflow.

## 2.3 Breast density masks generation using the proposed framework

In this study, Radiologist 1 annotated the central slice of each DBT study by creating a polygon mask and selecting an appropriate threshold value for that slice. Using our proposed method (section 2.2), we then generated a comprehensive density mask for each DBT image (176 volumes). An independent reader, Radiologist 2, segmented the same images using the same proposed approach. Additionally, Radiologist 2 visually inspected the DBT volumes and provided a visual assessment of volumetric breast percent density (referred to as Visual Assessment of Density, or VAD) (Table 1). Note that BI-RADS breast density is not available in the DBTex dataset.

| Visual Assessment of Density (VAD) | Number of Cases |
|---|---|
| 0-25 | 11 |
| 26-50 | 24 |
| 51-75 | 7 |
| 76-100 | 2 |
| Total | 44 |

**Table 1.** Distribution of cases across Visual Assessment of Density (VAD) categories used in the analysis.

---

[1] https://github.com/V-kr0pt/density_segmentation_gui

## 2.4 Evaluation Strategies

We evaluated two aspects of our method. The first was about the feasibility of using the proposed method for ground truth mask generation. We therefore evaluated whether two users can create similar dense tissue masks using our proposed approach (precision). Specifically, we assessed inter-reader agreement by computing patient-wise Dice similarity coefficients between dense tissue masks generated independently by two radiologists using the proposed approach. For each patient, four Dice coefficients were calculated, one per view (RCC, LCC, RMLO, and LMLO), by comparing the segmentations generated by the two users. The coefficients were subsequently averaged to yield a single Dice score per patient.

The second was about the accuracy of our time- and labor- saving algorithm. For this, we assessed the agreement between dense tissue masks by our approach compared to a radiologist's manual segmentation on a selected non-central slices. Specifically, we assessed the accuracy of the proposed method by computing the Dice score between proposed method (from Radiologist 2) and manual segmentations (from Radiologist 1) on the 20th and 80th percentile slices of the selected DBT volumes from 44 patients. The 20th and 80th percentiles were selected because they are sufficiently distant from the central slice while avoiding the peripheral regions of the volume, where increased blurring and less relevant information are present. We selected one CC and one MLO view per patient, randomly choosing between right and left views, for a total of 176 slices. Patient-wise Dice similarity coefficients were computed between the manual annotations and the corresponding masks generated by the proposed approach, providing practical validation against time-consuming manual slice-by-slice segmentation.

## Results

Figure 2 presents a bar plot of patient-wise Dice scores between the masks generated by two radiologists across different VAD groups. The proposed approach demonstrated strong inter-reader agreement, with a median patient-wise Dice similarity coefficient of 0.84 (standard deviation, SD = 0.12) between the two radiologists. We found no systematic difference between CC and MLO views (Dice CC = 0.85; SD = 0.13 vs. Dice MLO = 0.82; SD = 0.13). We found a higher median Dice value of 0.89 (SD = 0.14) for DBT volumes with VAD values higher than 50%, compared to those DBT volumes with VAD values below 51% (Median Dice = 0.81; SD = 0.13). This is expected, as there is a smaller number of pixels with dense tissue; therefore, a small difference between two masks can reduce the Dice values significantly.

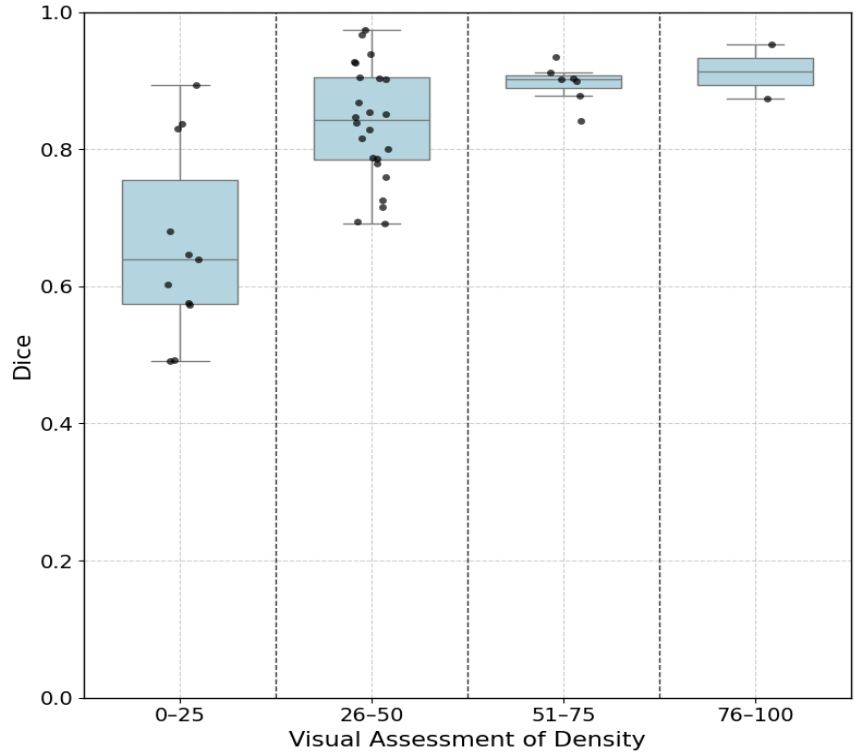

**Figure 2.** Boxplots showing patient-level Dice values stratified by Visual Assessment of Density (VAD). Boxes represent the interquartile range (Q1-Q3; IQR), the horizontal line within each box indicates the median, and overlaid points correspond to individual case Dice values. The whiskers show 1.5 x IQR.

For slice-based validation (Figure 3) against manual reference segmentations (20th and 80th percentile slices), the Dice score was 0.83 (SD = 0.11), which shows the accuracy and agreement of the generated masks with manual segmentation across selected DBT slices. For 20th percentile slices the Dice score is 0.83 (SD = 0.11) for 80th percentile slices the Dice score is 0.84 (SD = 0.11).

Figure 4 shows sample results across four different VAD values from two radiologists (Rad 1 and Rad 2). The figure demonstrates visual agreement between the resulting dense tissue masks.

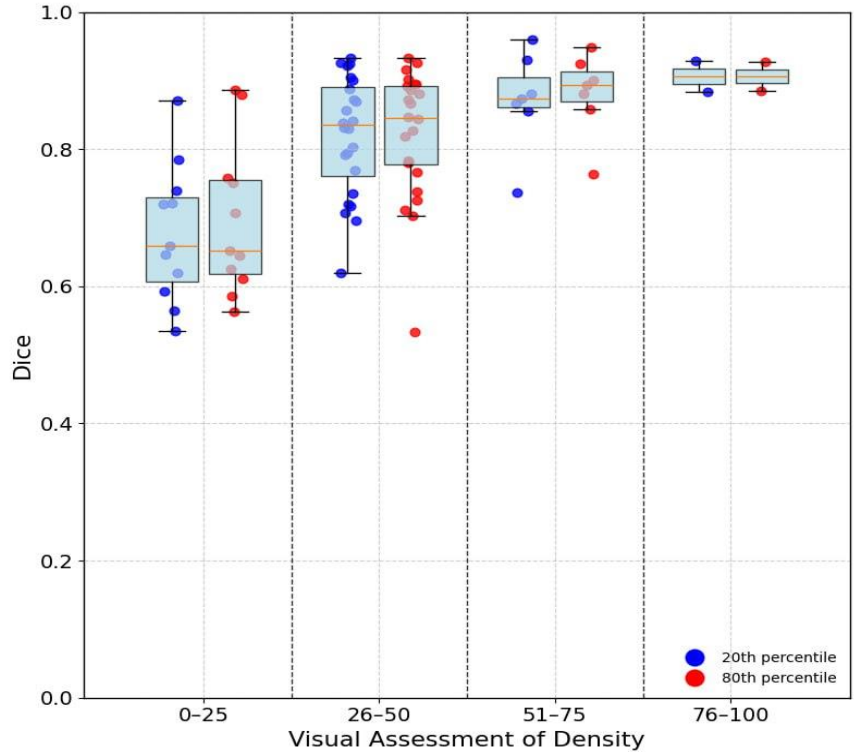

**Figure 3.** Boxplots showing patient-level Dice values comparing manual segmentations and the proposed method at the 20th and 80th slice indices, stratified by Visual Assessment of Density (VAD). Boxes represent the interquartile range (Q1–Q3; IQR), the horizontal line within each box indicates the median, and overlaid points correspond to individual case Dice values. The whiskers show 1.5 x IQR.

## Discussion

Our results demonstrate good inter-user agreement, with a median Dice score of 0.84 (SD = 0.12), highlighting the robustness of the method across users showing the feasibility of our proposed method of being used by a group of experts (high precision). In the second analysis, the proposed method demonstrated good accuracy with a median Dice score of 0.83 (SD = 0.11) when its segmentations were compared with manually segmented 20th and 80th percentile slices used as the reference standard.

We then conducted post-hoc analysis on outliers with lower Dice values between two expert-generated masks. Figure 5 shows an example with a VAD value < 25. This discrepancy resulted from different threshold values selected by the two users for a single volume from the same patient. Specifically, the outlier corresponds to a single LMLO volume with a Dice score of 0.14, whereas the other views (volumes) from the same patient (LCC, RMLO, and RCC) demonstrated higher Dice scores of 0.46, 0.76, and 0.73, respectively. This outlier is attributed to a user-related error and the complexity of the case due to predominantly fatty breast composition, as the other views from the same patient show consistently high Dice scores and therefore the agreement.

Our approach has several limitations. First, it is optimized for images acquired from the Hologic vendor (with the narrow scanning angle), which accounts for approximately 70% of the U.S. market [6]. Our approach may be suboptimal on images from other vendors, specifically with wider scanning angles. We will evaluate our approach on the other vendors including GE and Siemens as future study.

Second, the thresholding algorithm and the Dice calculation are sensitive to breast density amount. The most challenging cases for segmentation were those with fatty breasts, which contain minimal fibroglandular tissue and a relatively larger proportion of background structures such as vessels and skin folds. Because these structures can exhibit pixel intensities similar to fibroglandular tissue, they may occasionally be included in the segmented masks. However, in fatty breasts the absolute amount of fibroglandular tissue is small, and vessels and skin folds typically occupy only a limited number of pixels due to the second step of our approach (Figure 1b). Therefore, their inclusion is unlikely to substantially affect downstream quantitative analyses such as density estimation. Additionally, the Dice similarity coefficient is inherently sensitive to the size of segmented objects. In cases with very small amounts of fibroglandular tissue, even minor pixel-level discrepancies can lead to relatively larger reductions in Dice scores.

## Conclusion

We introduced and validated a time- and labor-saving workflow that streamlines the generation of density segmentation masks for fibroglandular tissue in DBT volumes. This workflow provides a practical foundation for a method to create a complete DBT density mask generated by a radiologist. This dataset can be used as "truth" to support the development of specialized deep learning algorithms for dense tissue segmentation in DBT volumes and to facilitate future research in breast cancer risk prediction.

## Acknowledgements:

This work (or closely related research) has not been published or accepted for publication, and it is not under consideration for publication. This manuscript is based on scientific content previously reported in the proceeding paper of the SPIE IWBI conference 2024 [7]. The authors declare there are no financial interests, commercial affiliations, or other potential conflicts of interest that have influenced the objectivity of this research or the writing of this paper. The GitHub code [8]
Data Availability: The data that support the findings of this study are available from the corresponding author, TM, upon reasonable request.

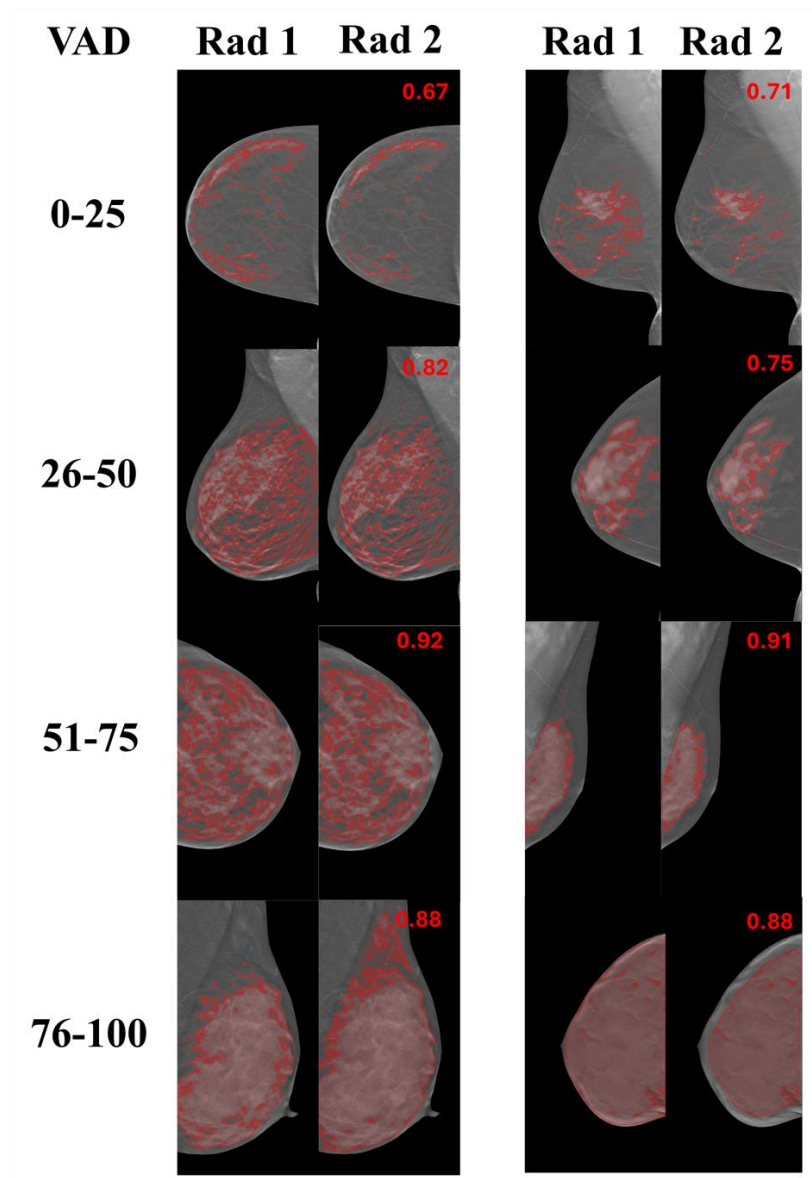

**Figure 4.** Example segmentation results from two radiologists for 8 different patients are shown. The figure demonstrates strong agreement across four different VAD values. For consistency, the 20th slice from each volume is displayed. The volume-wise Dice scores for each pair are shown in the upper right corner.

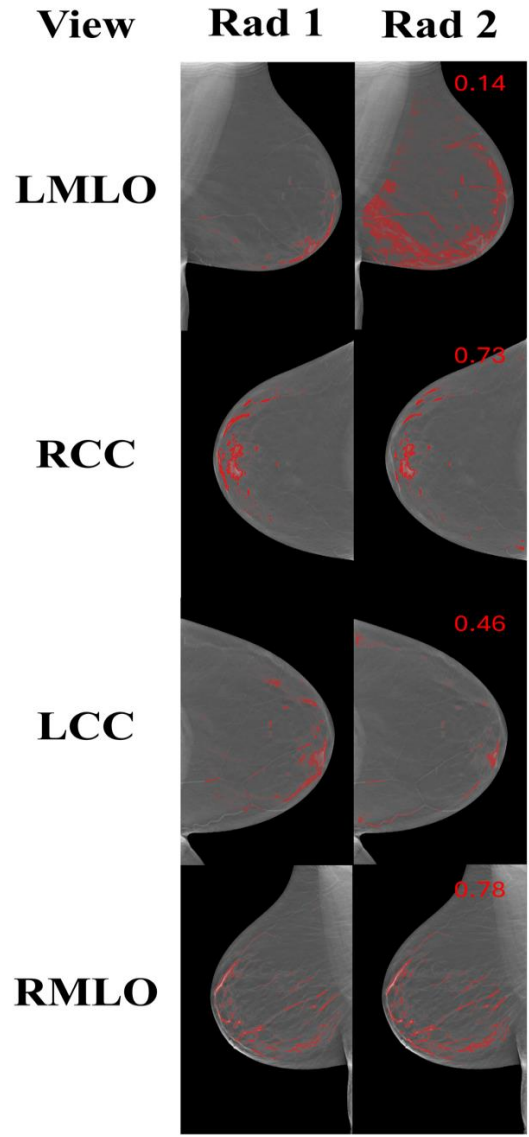

**Figure 5.** Shows outlier case from a single patient (VAD < 25%). For consistency, the 20th slice from each volume is displayed. The volume-wise Dice values for each pair are shown in the upper right corner. The Dice value for the LMLO view was lower than for the other views from the same patient. Furthermore, Radiologist 2 segmented a greater amount of tissue in the LMLO view compared to the other views, suggesting an error by Radiologist 2, and this error propagated through the slices, which explains the low Dice score.